\documentclass[10pt,twocolumn,letterpaper]{article}

\usepackage{cvpr}              % To produce the CAMERA-READY version
\usepackage[dvipsnames]{xcolor}

\definecolor{cvprblue}{rgb}{0.21,0.49,0.74}
\usepackage[pagebackref,breaklinks,colorlinks,citecolor=cvprblue]{hyperref}
\usepackage{graphicx}
\usepackage{amsmath}
\usepackage{amssymb}
\usepackage{booktabs}
\usepackage{hyperref}       % hyperlinks
\usepackage{url}            % simple URL typesetting
\usepackage{amsfonts}       % blackboard math symbols
\usepackage{nicefrac}       % compact symbols for 1/2, etc.
\usepackage{microtype}      % microtypography
\usepackage{xcolor}         % colors
\usepackage{adjustbox}
\usepackage{multicol}
\usepackage{multirow}
\usepackage{utfsym}
\usepackage{algpseudocode}
\usepackage{algorithm}
\usepackage{enumitem}
\usepackage{wrapfig}
\usepackage{caption}
\usepackage{amsmath}
\usepackage{amssymb}
\usepackage{dsfont}
\usepackage{pifont}
\usepackage[capitalize]{cleveref}
\crefname{section}{Sec.}{Secs.}
\Crefname{section}{Section}{Sections}
\Crefname{table}{Table}{Tables}
\crefname{table}{Tab.}{Tabs.}

\usepackage[accsupp]{axessibility}

\def\paperID{10030} % *** Enter the Paper ID here
\def\confName{CVPR}
\def\confYear{2024}

\title{TASeg: Temporal Aggregation Network for LiDAR Semantic Segmentation}

\author{
    Xiaopei Wu\textsuperscript{\rm 1, 2}, 
    Yuenan Hou\textsuperscript{\rm 2$*$},
    Xiaoshui Huang\textsuperscript{\rm 2$*$},
    Binbin Lin\textsuperscript{\rm 3, 4},
    Tong He\textsuperscript{\rm 2}, \\
    Xinge Zhu\textsuperscript{\rm 5},
    Yuexin Ma\textsuperscript{\rm 6},
    Boxi Wu\textsuperscript{\rm 4},
    Haifeng Liu\textsuperscript{\rm 1}$\thanks{~Corresponding authors.}$,
    Deng Cai\textsuperscript{\rm 1}, 
    Wanli Ouyang\textsuperscript{\rm 2}\\
    {\small
    \textsuperscript{\rm 1}State Key Lab of CAD\&CG, Zhejiang University \quad
    \textsuperscript{\rm 2}Shanghai AI Laboratory \quad
    \textsuperscript{\rm 3}Fullong Inc.}\\
    {\small
    \textsuperscript{\rm 4}School of Software Technology,     Zhejiang University \quad
    \textsuperscript{\rm 5}The Chinese University of Hong Kong  \quad
    \textsuperscript{\rm 6}ShanghaiTech University}
 }
\begin{document}
\maketitle
\def\algorithmname{TASeg}

\vspace{-2mm}
\begin{abstract}
% !TEX root = ../main.tex
Training deep models for LiDAR semantic segmentation is challenging due to the inherent sparsity of point clouds.
Utilizing temporal data is a natural remedy against the sparsity problem as it makes the input signal denser.
However, previous multi-frame fusion algorithms fall short in utilizing 
sufficient temporal information due to the memory constraint, and they also ignore the informative temporal images.
To fully exploit rich information hidden in long-term temporal point clouds and images, 
we present the Temporal Aggregation Network, termed \textbf{TASeg}.
Specifically, we propose a Temporal LiDAR Aggregation and Distillation (TLAD) algorithm, which 
leverages historical priors to assign different aggregation steps for different classes.
It can largely reduce memory and time overhead while achieving higher accuracy. 
Besides, TLAD trains a teacher injected with gt priors to distill the model, further boosting the performance.
To make full use of temporal images, we design a Temporal Image Aggregation and Fusion (TIAF) module, 
which can greatly expand the camera FOV and enhance the present features. 
Temporal LiDAR points in the camera FOV are used as mediums to transform temporal image features 
to the present coordinate for temporal multi-modal fusion.
Moreover, we develop a Static-Moving Switch Augmentation (SMSA) algorithm, 
which utilizes sufficient temporal information to enable objects to switch their motion states freely, 
thus greatly increasing static and moving training samples.
Our TASeg ranks \textbf{1$^{st}$}
\footnote[2]{~On the date of CVPR deadline, \textit{i.e.}, 2023-11-18 07:59 AM UTC.}
on three challenging tracks, \textit{i.e.}, SemanticKITTI single-scan track, multi-scan track and nuScenes LiDAR segmentation track, 
strongly demonstrating the superiority of our method. Codes are available at \url{https://github.com/LittlePey/TASeg}.
\vspace{-1mm}

\end{abstract}

\section{Introduction}
\label{sec:introduction}
% !TEX root = ../main.tex

LiDAR segmentation aims to infer the semantic information of each point in point clouds and plays an indispensable role in the autonomous driving~\cite{yurtsever2020survey, chen2023end, yolop, depthformer}. With the advent of deep learning, a large quantity of LiDAR segmentation algorithms~\cite{cylinder3d,randla-net,2dpass,sphereformer,uniseg} have been put forward.
% and have dominated the leaderboard of many popular benchmarks~\cite{semantickitti,nuscenes,waymo}.
% Despite the impressive performance exhibited by those methods,
Despite their impressive results,
the segmentation performance is still constrained by the inherent sparsity of point clouds.
To alleviate this, it is desirable to aggregate temporal data.
Previous multi-frame models \cite{temporallidarseg,svqnet,mars3d} can only fuse a few LiDAR frames due to the GPU memory constraint, which restricts them from utilizing rich information hidden in the long-term temporal point clouds. 
Although MSeg3D~\cite{mseg3d} stacks dozens of LiDAR frames, the memory and time overhead are intolerable. 
Moreover, previous attempts concentrate on utilizing temporal point clouds, ignoring the valuable information hidden in temporal images. 

To better leverage long-term temporal information, we propose a \textit{Temporal LiDAR Aggregation and Distillation} (TLAD) algorithm. It can perform efficient multi-frame aggregation while achieving higher accuracy. TLAD consists of \textit{Flexible Step Aggregation} (FSA) and \textit{Mask Distillation}. FSA is based on the observation that the model needs different amounts of temporal points for different classes.
For difficult classes, such as bicyclists, more points are needed to yield accurate predictions. However, for easy classes, such as cars and roads, it is unnecessary to aggregate too many point cloud frames for them. 
Therefore, we propose to assign different aggregation steps for different classes in point cloud sequences, which can significantly save memory and computation overhead on easy classes while providing sufficient temporal points for difficult classes.
To identify classes of temporal point clouds, historical predictions are used, considering the temporal data is processed in chronological order, which is consistent with practical applications.

Interestingly, by assigning different classes with different steps, FSA actually injects historical priors into the aggregated point clouds. For classes with different steps, their patterns can be more discriminative due to their different densities. This makes the point cloud frames accumulated by FSA more conducive for segmentation. Experimental results show that FSA can not only reduce memory and time costs but also enhance overall performance. To further verify our idea, we directly replace historical predictions of temporal point clouds with their ground truth labels for FSA. We find the performance boosts greatly. This inspires us to use the model injected with gt priors to distill the model injected with historical priors, which we term \textit{Mask Distillation}.

To fully use the informative temporal images, we devise a \textit{Temporal Image Aggregation and Fusion} (TIAF) module. Previous multi-modal fusion methods usually suffer from limited image features due to the different FOVs (field of views) between the LiDAR and the camera. We observe that as the ego-vehicle moves forward, cameras can capture different FOVs. By aggregating images of different timestamps, the FOV of the present camera can be enlarged greatly.
In addition, temporal images can provide different views for the same region at different timestamps, which can supply richer information.
Our TIAF leverages temporal LiDAR points in the FOV of the corresponding temporal camera to transform historical image features to the present coordinate with the pose matrix. After aggregating all temporal image features, we use a series of 3D sparse convolutions to fuse them and convert them to voxel representation. Ultimately, we use temporal LiDAR points to gather temporal image features for temporal multi-modal fusion. The fused features embrace both temporal LiDAR and temporal image information, which contribute to more precise segmentation results. To our knowledge, we are the first to leverage temporal point clouds and temporal images simultaneously for LiDAR semantic segmentation.

Moreover, in the multi-scan task, the model needs to distinguish the motion states (moving or static) of a movable object, which is very challenging. In this paper, we make full use of temporal information to remarkably increase the multi-scan perception ability of the model. Specifically, we propose a data augmentation strategy named \textit{Static-Moving Switch Augmentation} (SMSA). It manipulates temporal point clouds of a movable object to switch its motion state. In this way, we can increase the quantity of static and moving samples significantly, even in the absence of training samples of static or moving classes.

In summary, our major contributions are listed as follows:
\vspace{-2mm}
\begin{itemize}[leftmargin=*]
\item We propose a \textit{Temporal LiDAR Aggregation and Distillation} algorithm, which uses \textit{Flexible Step Aggregation} and \textit{Mask Distillation} techniques to largely reduce memory and time costs while achieving higher accuracy.

\vspace{2mm}
\item We devise a \textit{Temporal Image Aggregation and Fusion} module, which exploits temporal images to enlarge the camera FOV and enhance present features. It also delivers a scheme for temporal multi-modal fusion.

\vspace{2mm}
\item We design a \textit{Static-Moving Switch Augmentation} algorithm to enable static and moving objects to switch their motion states freely. With this technique, we can greatly increase static and moving training samples.

\item Our Temporal Aggregation Network, dubbed~\textbf{TASeg}, achieves impressive results on SemanticKITTI and nuScenes benchmarks. Notably, our TASeg \textit{ranks 1$^{st}$} on three challenging tracks. Thorough ablation studies are provided to demonstrate the efficacy of our approach.
\end{itemize}

\section{Related Work}
\label{sec:relatedwork}
% !TEX root = ../main.tex

\noindent \textbf{LiDAR Semantic Segmentation.} 
LiDAR semantic segmentation aims to assign a unique class label to each point in the input point cloud sequence. Recent years have witnessed an explosion of LiDAR segmentation algorithms~\cite{pointnet++,minkunet,af2s3net,spvnas,pvkd,2dpass,sphereformer,mseg3d,rangeformer,uniseg}. 
For example, \cite{pointnet} is the pioneering work that approximates the permutation-invariant function with a per-point and shared Multi-Layer Perceptron. 
\cite{cylinder3d} changes traditional cubic grids to cylindrical grids and designs a network of asymmetrical 3D convolutions. \cite{sphereformer} divides the space with the radial window, which increases the receptive field smoothly and helps improve the performance. 
Despite their good segmentation performance, these methods still take the single LiDAR frame as input, which does not utilize the rich semantic and geometric information hidden in temporal data.

\vspace{3mm}
\noindent \textbf{Multi-Frame LiDAR Perception.}
Compared to a single LiDAR scan, multiple LiDAR scans can provide more sufficient information. Recently, many research efforts have been put on temporal LiDAR segmentation \cite{4d-pls,4d-stop,eq-4d, SpSequenceNet,4d_minknet,ASAP-Net, temporallidarseg, mars3d, deeptemporalseg, mseg3d}. 
For example, \cite{deeptemporalseg} leverages a Bayes filter to explore the temporal consistency.
\cite{svqnet} shunts the historical points into two groups to utilize historical frames efficiently.
Despite the success of previous multi-frame methods, they can not leverage the valuable information hidden in long-term temporal point clouds due to the GPU memory constraint. Although \cite{mseg3d} stacks dozens of LiDAR scans, the memory and time overhead are intolerable. In this paper, we present an efficient multi-frame aggregation algorithm, which can greatly save memory and computation consumption while achieving higher performance.

\vspace{3mm}
\noindent \textbf{Multi-Modal Fusion.}
Since LiDAR and camera are two complementary sensors for 3D semantic segmentation, multi-modal fusion has gained increasing attention in recent years \cite{pmf,el2019rgb,fuseseg,lcps,uniseg}.
However, these multi-modal fusion methods usually suffer from limited overlapped regions between the LiDAR and camera due to their different FOVs. \cite{2dpass} proposes a cross-modal knowledge distillation method, which is free from images at inference, while it causes much loss of the RGB information. \cite{mseg3d} completes the missing camera features using predicted pseudo-camera features, while the used image information is still restricted to the present camera FOV. In this paper, we fully take advantage of temporal images to enlarge the FOV of the present camera and enhance present image features.

\noindent \textbf{Knowledge Distillation for LiDAR Perception.}
Knowledge distillation \cite{kd} is widely used in various fields in that it can improve the performance of the student without sacrificing inference efficiency.
In LiDAR perception, \cite{sparsekd, pvkd} compress a cumbersome teacher to a lightweight student to reinforce the representation learning of the student as well as maintain high efficiency.
\cite{2dpass} distills the prior of 2D images to 3D point clouds with well-designed cross-modal knowledge distillation module.
\cite{sparse2dense,smf-ssd} utilizes a multi-frame teacher to help a single-frame student learn dense 3D features.
By contrast, our method aims to transfer knowledge from a multi-frame teacher injected with gt priors to a multi-frame student injected with historical priors.

\begin{figure*}[t]
	\begin{center}
		\setlength{\fboxrule}{0pt}
		\fbox{\includegraphics[width=0.98\textwidth]{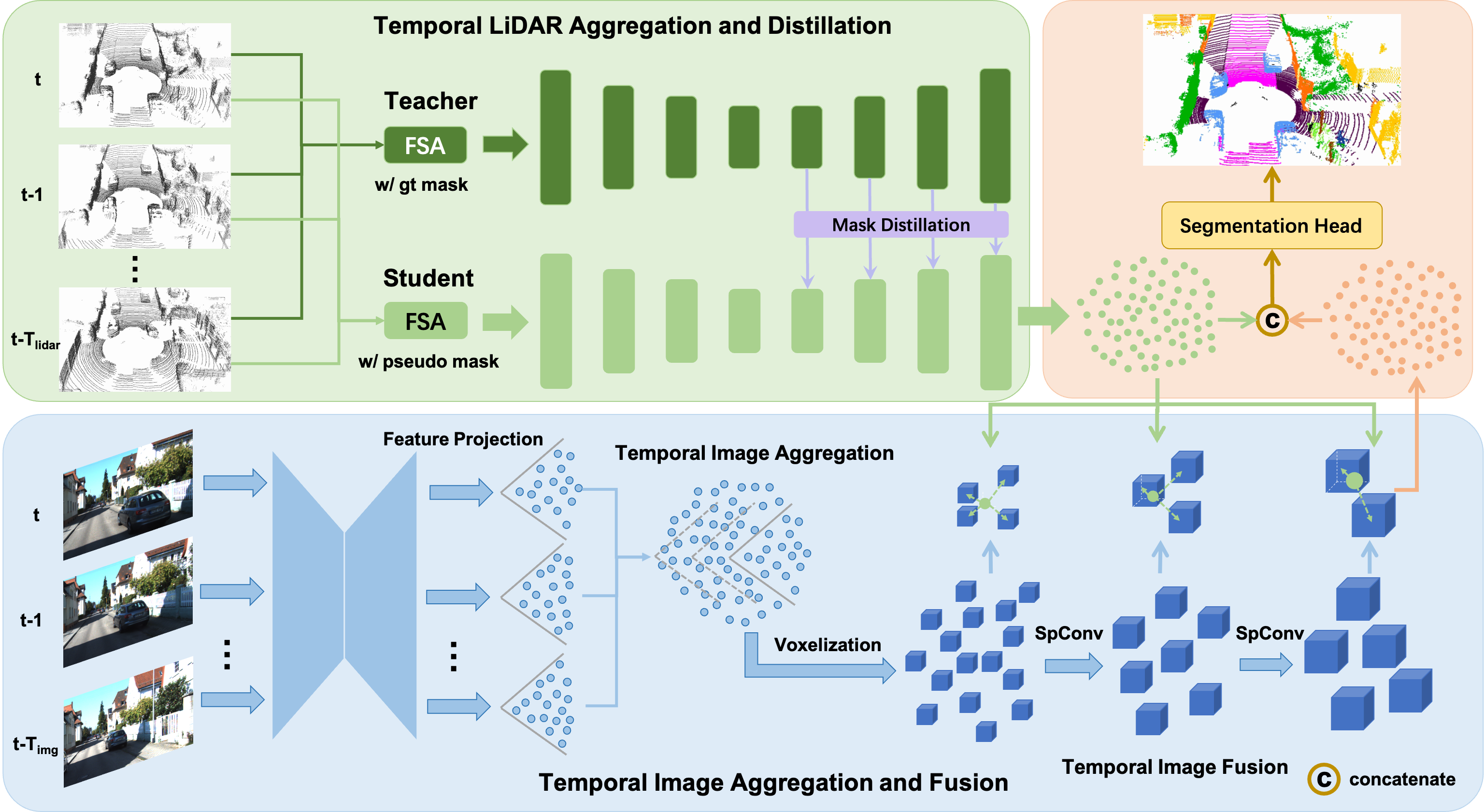}}
	\end{center}
	\vspace{-5mm}
	\caption{Overview of our Temporal Aggregation Network (TASeg).
(1) \textit{Temporal LiDAR Aggregation and Distillation} leverages the proposed Flexible Step Aggregation (FSA)
to assign different temporal steps for different classes, and it utilizes a teacher injected with gt priors for knowledge distillation. 
(2) \textit{Temporal Image Aggregation and Fusion} takes temporal 
LiDAR points as mediums to transform historical image 
features to the present coordinate. 3D sparse convolutions are employed to fuse temporal image features.
Finally, we use temporal LiDAR points to gather voxel-wise temporal image features for temporal multi-modal fusion.
}
	\label{fig:overview}
	\vspace{-5mm}
\end{figure*}

\section{Methodology}
\label{sec:methodology}
\subsection{Temporal LiDAR Aggregation and Distillation}
A simple solution to utilize the temporal information of a consecutive point cloud sequence is to concatenate all points of the sequence, as shown in the following equation:
\begin{equation}
\begin{aligned}
    X_{t} &= \text{concat}(P_{t}, T_{t-1}P_{t-1},  \dots, T_{t-\Delta t}P_{t-\Delta t}),\\
    Y_{t} &= \text{concat}(L_{t},  L_{t-1},  \dots,  L_{t-\Delta t}),\\
\end{aligned}
\end{equation}
where $P_{t-i}$ and $L_{t-i}$ denote the $(t-i)^{th}$ point cloud frame and the corresponding point-wise label. $T_{t-i}$ is the transformation matrix that transforms the coordinate from the $(t-i)^{th}$ frame to the $t^{th}$ frame. $\Delta t$ is the window size of temporal point clouds and $\text{concat}(.)$ denotes the concatenation operation. $X_{t}$ and $Y_{t}$ are the aggregated LiDAR frame and point-wise label.
Although simple, there are some problems. On the one hand, direct concatenation consumes much GPU memory. On the other hand, the huge memory cost constrains the multi-frame model from utilizing more historical frames, thereby limiting the ultimate performance.

\subsubsection{Flexible Step Aggregation}
To reduce the GPU memory consumption, we can sample temporal frames with a step, while this method is also sub-optimal. A small step introduces huge memory overhead, whilst a large step cannot utilize sufficient temporal information. 
To this end, we propose Flexible Step Aggregation (FSA). Our method is based on the observation that for different classes, the model needs different amounts of temporal information. For difficult classes, such as bicycles, more points are needed to yield accurate predictions. For easy classes, such as cars and buildings, it is unnecessary to aggregate many point cloud frames for them. 
Based on the above analysis, we propose to assign different steps for different classes according to their learning difficulty. Specifically, we leverage historical predictions to divide the temporal point clouds into several class groups and assign a specific sampling step for each group. Then, we aggregate temporal points for each group with the corresponding step. Eventually, temporal points of all groups are concatenated with the current frame, resulting in an aggregated frame. 

The group division is not strict as long as it follows the principle that more difficult classes need smaller steps. For example, we can simply divide all classes 
into three groups according to their segmentation performance, such as [0, 80) mIoU, [80, 90) mIoU and [90, 100] mIoU. Then, we assign the three groups with a step of 2, 4 and $\infty$ ($\infty$ means we do not aggregate temporal points for the group). 
To save more memory and computation without sacrificing performance, a more fine-grained division can be used. 
Formally, suppose we divide temporal point clouds into $g$ groups. We aggregate temporal points for the $k^{th}$ group as below:
\vspace{-1mm}
\begin{equation}
\begin{aligned}
    X_{t}^{k} &= \text{concat}( T_{o_1}P_{o_1}M_{o_1}^k, \dots, T_{o_n}P_{o_n}M_{o_n}^k),\\
    Y_{t}^{k} &= \text{concat}(L_{o_1}M_{o_1}^k,  \dots,  L_{o_n}M_{o_n}^k).\\
\end{aligned}
\vspace{-1mm}
\end{equation}
Here $o=\{o_i|o_i=t-i \times s_k, i=1,2,...,n, n=\lfloor\Delta t/s_k\rfloor\}$. $s_k$ is the sampling step for the $k^{th}$ class group and $\lfloor . \rfloor$ is the floor operation. $M^k_{o_i}$ is the group mask that indicates which point of $P_{o_i}$ belongs to the $k^{th}$ class group. It can be obtained from historical predictions. Finally, we concatenate points and labels of all groups with the current frame:
\vspace{-1mm}
\begin{equation}
\begin{aligned}
    X_{t} &= \text{concat}(P_{t}, X_{t}^{1}, \dots, X_{t}^{k}, \dots, X_{t}^{g}),\\
    Y_{t} &= \text{concat}(L_{t}, Y_{t}^{1}, \dots, X_{t}^{k}, \dots, Y_{t}^{g}).
\end{aligned}
\vspace{-1mm}
\end{equation}

In this way, we discard massive redundant temporal points while maintaining essential temporal information. 
Since easy classes are usually large and hold a large quantity of points, the GPU memory overhead can be further reduced.
Moreover, by assigning different classes with different steps, FSA actually injects historical priors into the aggregated point clouds. 
For classes with different steps, their patterns can be more discriminative due to their different densities (steps), 
which makes it easier to segment the multi-frame point clouds aggregated by FSA. 
Experiments in Table \ref{tab:kitti_fsa} verify that the proposed FSA can not only save memory and time costs but also achieve better performance.

\vspace{-2mm}
\subsubsection{Mask Distillation}
In FSA, we use historical predictions to generate group masks, which we call pseudo masks.
Since ground truth labels are more accurate than historical predictions,
a natural question arises: what if we use gt masks (the group masks generated by ground truth labels) for FSA?
Our experiment shows that the performance can be improved greatly.
Actually, using gt masks for FSA can produce more discriminative patterns for classes with different steps.
This motivates us to simulate features of a model trained on temporal point clouds aggregated with gt masks.
In this way, our model is guided to learn more discriminative features to distinguish different classes. We call this \textit{Mask Distillation}.
Specifically, we formulate the distillation on the feature map of student $F^s \in \mathbb{R}^{N^s \times C}$ and 
the feature map of teacher $F^t \in \mathbb{R}^{N^t \times C}$:
\begin{equation}
\begin{aligned}
\mathcal{L}_{\text{KD}} = \mathbb{E}[\| F^s m^s - F^t m^t \|_2].
\end{aligned}
\end{equation}
Since temporal point clouds aggregated with pseudo masks and gt masks are different, we use masks $m^s$ and $m^t$ to select voxels that appeared in both $F^s$ and $F^t$. Note that our Mask Distillation is distinct from the methods that distill a multi-frame model to a single-frame model~\cite{sparse2dense, smf-ssd}. Both the student and teacher in our method are multi-frame models.

\subsection{Temporal Image Aggregation and Fusion}
Previous multi-modal fusion methods only focus on leveraging present images while ignoring the precious value of temporal images.
Temporal images can provide broader camera FOVs and richer information.
Besides, they can enable more robust multi-modal fusion under the malfunction condition on some cameras.
In this section, we provide an effective solution for aggregating temporal image features and performing temporal multi-modal fusion.

\textit{Temporal Aggregation and Fusion.} Since temporal images are in different feature spaces, it is difficult to 
establish the relationship between different images for feature aggregation.
In our method, we take temporal LiDAR points as mediums to transform temporal image features to the present coordinate with the pose information. 
This way, temporal image features are unified to the present 3D space.
Specifically, given an image $I_{t-\Delta t} \in \mathbb{R}^{H \times W \times 3}$ and point cloud $P_{t-\Delta t} \in \mathbb{R}^{N \times 3}$, 
we use an image network to extract the image feature $Z_{t-\Delta t} \in \mathbb{R}^{H \times W \times C}$. 
According to the sensor calibration, we can establish the pixel-to-point mapping between 2D pixels and 3D points. 
Hence, we can project image feature $Z_{t-\Delta t}$ to 3D space, resulting in point-wise image feature $Q_{t-\Delta t} \in \mathbb{R}^{M \times C}$,
where $M$ is the number of LiDAR points located on $Z_{t-\Delta t}$. 
By transforming $Q_{t-\Delta t}$ to the present coordinate with the pose matrix, 
we realize the aggregation of temporal image features:
\vspace{-2mm}
\begin{equation}
\begin{aligned}
    \overline{X}_{t} &= \text{concat}(Q_{t}, T_{o_1}Q_{o_1}, \dots, T_{o_n}Q_{o_n}),\\
\end{aligned}
\vspace{-1mm}
\end{equation}
where $o=\{o_i|o_i=t-i \times s, i=1,2,...,\lfloor \Delta t/s\rfloor\}$ and $s$ is the sampling step for temporal images.
With temporal images aggregated, the FOV of the present camera is expanded greatly. 
Moreover, temporal image feature fusion becomes convenient because they are 
unified to the same 3D space. Concretely, we can use several 3D sparse convolutions 
to fuse aggregated temporal image features, which also endows them with geometric information, as shown in Equation \ref{eq:voxelization}. In addition, feature map downsampling is also utilized to generate multi-scale voxel features, providing richer information for subsequent temporal multi-modal fusion.

\begin{equation}
\begin{aligned}
    \overline{V}_{t} &= \text{SparseConv}_{\text{3D}}(\text{Voxelization}(\overline{X}_{t})).\\
\end{aligned}
\label{eq:voxelization}
\end{equation}

\textit{Temporal Multi-Modal Fusion.}
Benefiting from the temporal image aggregation, associating temporal image features with temporal LiDAR points also becomes convenient. 
Specifically, given temporal image features converted to unified voxel representation $\overline{V}_t$, 
we can establish a point-to-voxel association between temporal LiDAR points and $\overline{V}_t$.
For each temporal LiDAR point, we generate its image feature by pooling its nearby 
voxel-wise temporal image features with trilinear interpolation instead of hard indexing. 
To extract richer image features, we perform interpolation on multi-scale feature maps.
Finally, we concatenate the point cloud features and aggregated multi-scale image features, 
resulting in fused features, which convey powerful information of both temporal point clouds and temporal images.

\vspace{2mm}
\textit{2D and 3D Supervision.}
To make the extracted image features more informative, we add 2D supervision and 3D supervision on the 2D backbone and 3D convolutions in the image branch, respectively. The 3D supervision is just the label of point clouds.
The 2D supervision is obtained by projecting labels of point clouds to the image plane.

\subsection{Static-Moving Switch Augmentation}
In the multi-scan task, the model is required to distinguish the motion state of movable objects. 
To enable the model to explore a large data space, 
we design an effective data augmentation, Static-Moving Switch Augmentation (SMSA).
SMSA enables a movable object to switch its motion state freely, which can remarkably increase the sample quantity of static and moving objects.
Concretely, considering that a unique object $b$ has the same instance id in all frames of a sequence, 
we can use its instance mask to crop its temporal point clouds, which is denoted as $\mathcal{B}$ = $\{\mathcal{B}_{o_i} | i=0,1,...,\lfloor \Delta t/s\rfloor, \mathcal{B}_{o_i} \in \mathbb{R}^{N \times 3}\}$. 
Here $o=\{o_i|o_i=t-i \times s, i=1,2,...,\lfloor \Delta t/s\rfloor\}$, $s$ is the sampling step and $\mathcal{B}_{o_i}$ is the temporal part of $b$ at $t_{o_i}$.
By manipulating $\mathcal{B}$, we can change the motion state of $b$. 

\begin{figure}[t]
	\begin{center}
		\setlength{\fboxrule}{0pt}
		\fbox{\includegraphics[width=0.48\textwidth]{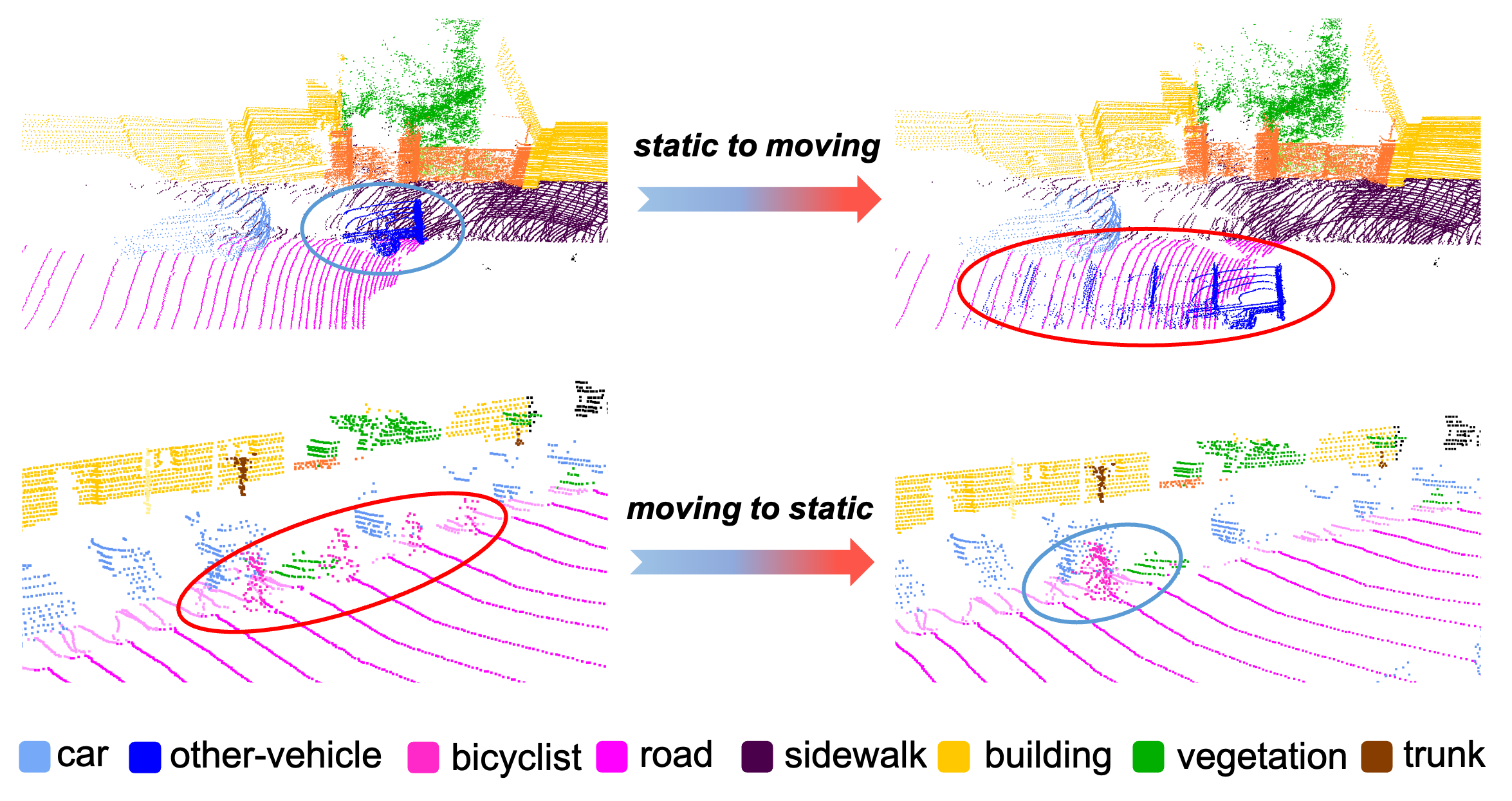}}
	\end{center}
	\vspace{-6mm}
	\caption{Visualization of the augmented samples by our Static-Moving Switch Augmentation (SMSA). Our SMSA switches the motion state by manipulating the temporal parts of objects.}
	\label{fig:smsa}
	\vspace{-5mm}
\end{figure}

\vspace{2mm}
\textit{Static to Moving.}
If $b$ is static, all temporal parts of $b$ locate at the same position, 
as shown in the upper-left of Figure \ref{fig:smsa}.
To change $b$ to a moving object, we can shift each temporal part of $b$ with an offset.
Considering that objects typically move at a constant speed within a short time, we set the offset between adjacent temporal parts to be the same. 
For offset itself, it is a random value to increase the diversity of resulting moving samples.
As for the direction of the offset, it can be roughly estimated by comparing the 
width and height of the minimum bounding box of $b$. 
Since static objects often park on the side of the road, which is crowded,
the shifted temporal parts of static objects may overlap with other objects.
To alleviate this, we define a set of anchor points around $b$ and 
a coverage area for each anchor point. 
Then, we shift all temporal parts of $b$ to the anchor point whose coverage area contains the fewest LiDAR points (refer to the supplementary material for more details).

\vspace{1mm}
\textit{Moving to Static.}
If $b$ is moving, different temporal parts of $b$ locate at different positions.
To switch its motion state to static, we can shift all its temporal parts to the same position.
In particular, we calculate the centers of each temporal part of $b$, 
which are denoted as $\mathcal{C}$ = $\{\mathcal{C}_{o_i}| \mathcal{C}_{o_i} \in \mathbb{R}^{3}\}$.
Considering the trajectory of moving objects in a short time is approximate to a line 
and the speed is a constant, we simply use $q = \mathcal{C}_{o_0} - \mathcal{C}_{o_1}$ as 
the offset of adjacent temporal parts of $b$. Eventually, we can obtain a static object 
by shifting all temporal parts of $b$ to ${C}_{o_0}$ (see Figure \ref{fig:smsa}).

\vspace{1mm}
It should be noted that our SMSA is an online and plug-and-play data augmentation strategy, 
which consumes negligible storage and computation costs.

\subsection{Training Objective}
The final training objective is comprised of LiDAR segmentation loss, mask distillation loss, 
fusion segmentation loss, 2D and 3D supervision loss on the image branch:
\vspace{-2mm}
\begin{equation}
\begin{aligned}
\mathcal{L} = \mathcal{L}_{\text{LiDAR}} + 
\alpha \mathcal{L}_{\text{KD}} + 
\beta \mathcal{L}_{\text{Fusion}} + 
\gamma (\mathcal{L}_{\text{2D}} + \mathcal{L}_{\text{3D}}),
\vspace{-2mm}
\end{aligned}
\end{equation}
where $\alpha, \beta, \gamma$ are the coefficients to control the effect of different losses. We set $\alpha=1, \beta=1, \gamma=1$ by default.

\section{Experiments}
\label{sec:experiments}
% !TEX root = ../main.tex
\begin{table*}[t]
    \renewcommand\tabcolsep{1.5pt}
    \begin{center}
        {\fontfamily{cmr}
            \resizebox{\textwidth}{!}{
                \begin{tabular}{lc|ccccccccccccccccccc}
\hline
Method&
\rotatebox{90}{mIoU}& \rotatebox{90}{car}& \rotatebox{90}{bicycle}& \rotatebox{90}{motorcycle}&
\rotatebox{90}{truck}& \rotatebox{90}{other-veh.}& \rotatebox{90}{person}& \rotatebox{90}{bicyclist}&
\rotatebox{90}{motorcyclist~}& \rotatebox{90}{road}& \rotatebox{90}{parking}& \rotatebox{90}{sidewalk}&
\rotatebox{90}{other-gro.}& \rotatebox{90}{building}& \rotatebox{90}{fence}& \rotatebox{90}{vegetation}&
\rotatebox{90}{trunk}& \rotatebox{90}{terrain}& \rotatebox{90}{pole}& \rotatebox{90}{traffic sign}\\
\hline
\hline
SqueezeSegV2~\cite{squeezesegv2} & 39.7& 81.8 & 18.5 & 17.9 & 13.4 & 14.0 & 20.1 & 25.1 & 3.9 & 88.6 & 45.8 & 67.6 & 17.7 & 73.7 & 41.1 & 71.8 & 35.8 & 60.2 & 20.2 & 26.3  \\
RangeNet53++~\cite{rangenet++} & 52.2& 91.4 & 25.7 & 34.4 & 25.7 & 23.0 & 38.3 & 38.8 & 4.8 & 91.8 & 65.0 & 75.2 & 27.8 & 87.4 & 58.6 & 80.5 & 55.1 & 64.6 & 47.9 & 55.9  \\
RandLA-Net~\cite{randla-net} & 55.9& 94.2 & 29.8 & 32.2 & 43.9 & 39.1 & 48.4 & 47.4 & 9.4 & 90.5 & 61.8 & 74.0 & 24.5 & 89.7 & 60.4 & 83.8 & 63.6 & 68.6 & 51.0 & 50.7  \\
SqueezeSegV3~\cite{squeezesegv3} & 55.9& 92.5 & 38.7 & 36.5 & 29.6 & 33.0 & 45.6 & 46.2 & 20.1 & 91.7 & 63.4 & 74.8 & 26.4 & 89.0 & 59.4 & 82.0 & 58.7 & 65.4 & 49.6 & 58.9  \\
JS3C-Net~\cite{js3c-net} & 66.0& 95.8 & 59.3 & 52.9 & 54.3 & 46.0 & 69.5 & 65.4 & 39.9 & 88.9 & 61.9 & 72.1 & 31.9 & 92.5 & 70.8 & 84.5 & 69.8 & 67.9 & 60.7 & 68.7  \\
SPVNAS~\cite{spvnas} & 67.0& 97.2 & 50.6 & 50.4 & 56.6 & 58.0 & 67.4 & 67.1 & 50.3 & 90.2 & 67.6 & 75.4 & 21.8 & 91.6 & 66.9 & 86.1 & 73.4 & 71.0 & 64.3 & 67.3  \\
Cylinder3D~\cite{cylinder3d} & 68.9& 97.1 & 67.6 & 63.8 & 50.8 & 58.5 & 73.7 & 69.2 & 48.0 & 92.2 & 65.0 & 77.0 & 32.3 & 90.7 & 66.5 & 85.6 & 72.5 & 69.8 & 62.4 & 66.2  \\
RPVNet~\cite{rpvnet} & 70.3& 97.6 & 68.4 & 68.7 & 44.2 & 61.1 & 75.9 & 74.4 & 43.4 & 93.4 & 70.3 & 80.7 & 33.3 & 93.5 & 72.1 & 86.5 & 75.1 & 71.7 & 64.8 & 61.4  \\
(AF)$^2$-S3Net~\cite{af2s3net} & 70.8& 94.3 & 63.0 & 81.4 & 40.2 & 40.0 & 76.4 & 81.7 & 77.7 & 92.0 & 66.8 & 76.2 & 45.8 & 92.5 & 69.6 & 78.6 & 68.0 & 63.1 & 64.0 & 73.3  \\
PVKD~\cite{pvkd} & 71.2& 97.0 & 69.3 & 53.5 & 67.9 & 60.2 & 75.1 & 73.5 & 50.5 & 91.8 & 77.5 & 70.9 & 41.0 & 92.4 & 69.4 & 86.5 & 73.8 & 71.9 & 64.9 & 65.8  \\
2DPASS~\cite{2dpass} & 72.9& 97.0 & 63.6 & 63.4 & 61.1 & 61.5 & 77.9 & 81.3 & 74.1 & 89.7 & 67.4 & 74.7 & 40.0 & 93.5 & 72.9 & 86.2 & 73.9 & 71.0 & 65.0 & 70.4  \\
SphereFormer~\cite{sphereformer} & 74.8& 97.5 & 70.1 & 70.5 & 59.6 & 67.7 & 79.0 & 80.4 & 75.3 & 91.8 & 69.7 & 78.2 & 41.3 & 93.8 & 72.8 & 86.7 & 75.1 & 72.4 & 66.8 & 72.9  \\
UniSeg~\cite{uniseg} & 75.2 & 97.9 & 71.9 & 75.2 & 63.6 & 74.1 & 78.9 & 74.8 & 60.6 & 92.6 & 74.0 & 79.5 & 46.1 & 93.4 & 72.7 & 87.5 & 76.3 & 73.1 & 68.3 & 68.5 \\\hline
TASeg (Ours) & \textbf{76.5} & 97.7 & 71.8 & 71.4 & 65.2 & 78.7 & 79.9 & 84.6 & 78.6 & 91.6 & 74.0 & 78.0 & 39.3 & 93.5 & 73.4 & 86.6 & 75.0 & 71.7 & 69.6 & 73.8 \\\hline
                \end{tabular}
        }}
    \end{center}
    \vspace{-4mm}
    \caption{Comparison with state-of-the-arts on \textit{SemanticKITTI} test set (single-scan). All results can be found on the online leaderboard.}
    \vspace{-2mm}
    \label{tab:kitti_test}
\end{table*}

\begin{table*}[th]		
    \renewcommand\tabcolsep{1.5pt} 
    \begin{center}
        {\fontfamily{cmr}
            \resizebox{0.7\textwidth}{!}{%
                \begin{tabular}{lc|cccccccccccc}
                    \hline 
                    Method & 
                    \rotatebox{90}{mIoU}&
                    \rotatebox{90}{car-s}&
                    \rotatebox{90}{car-m}&
                    \rotatebox{90}{truc.-s}&
                    \rotatebox{90}{truc.-m}&
                    \rotatebox{90}{othe.-s}&
                    \rotatebox{90}{othe.-m~}&
                    \rotatebox{90}{pers.-s}&
                    \rotatebox{90}{pers.-m}&
                    \rotatebox{90}{bicy.-s}&
                    \rotatebox{90}{bicy.-m}&
                    \rotatebox{90}{moto.-s}&
                    \rotatebox{90}{moto.-m~} \\
                    \hline 
                    \hline
LatticeNet~\cite{latticenet} & 45.2 & 91.1  & 54.8  & 29.7  & 3.5   & 23.1  & 0.6   & 6.8   & 49.9  & 0.0   & 44.6  & 0.0   & 64.3  \\
TemporalLidarSeg~\cite{temporallidarseg} & 47.0 & 92.1  & 68.2  & 39.2  & 2.1   & 35.0  & 12.4  & 14.4  & 40.4  & 0.0   & 42.8  & 0.0   & 12.9  \\
(AF)$^2$-S3Net~\cite{af2s3net} & 56.9  & 91.8  & 65.3  & 15.7  & 5.6   & 27.5  & 3.9   & 16.4  & 67.6  & 15.1 & 66.4  & 67.1 & 59.6  \\
MarS3D~\cite{mars3d} & 52.7 &95.1 &78.4 &39.7 &5.1 &36.6 &10.0 &16.2 &58.0 &1.2 &67.3 &0.0 &36.3  \\
SVQNet~\cite{svqnet} & 60.5 &96.0 &80.1 &41.0 &5.1 &60.4 &7.1 &28.7 &85.1 &0.0 &72.4 &0.0 &88.1 \\
2DPASS~\cite{2dpass} &  62.4 & 96.2  & 82.1  & 48.2  & 16.1  & 52.7  & 3.8   & 35.4  & 80.3  & 7.9   & 71.2  & 62.0  & 73.1  \\
\hline
TASeg (Ours) & \textbf{65.7} & 94.8 & 72.8     & 53.6 & 35.2    & 71.4 & 53.2    & 36.7 & 80.4    & 32.4 & 75.2    & 12.3 & 89.6 \\
\hline
        \end{tabular}}}
    \end{center}
    \vspace{-4mm}
    \caption{Comparison to the state-of-the-art methods on \textit{SemanticKITTI} test set (multi-scan). \textit{-s} indicates static and \textit{-m} stands for moving.}
    \vspace{-4mm}
    \label{tab:kitti_multi_test}
\end{table*}

\noindent \textbf{Datasets \& Evaluation Metrics.} Following~\cite{cylinder3d,pvkd}, we evaluate the performance on two popular LiDAR segmentation benchmarks, \emph{i.e.}, SemanticKITTI and nuScenes. SemanticKITTI contains 22 point cloud sequences, where sequences 0-10, 8, 11-21 are selected as training, validation and testing, respectively. As for nuScenes, it collects 1, 000 driving scenes, where 850 scenes are chosen for training and validation, and the remaining 150 scenes are used for testing. We conduct experiments on three tracks, \emph{i.e.}, SemanticKITTI single-scan and multi-scan semantic segmentation and nuScenes semantic segmentation.

\noindent \textbf{Implementation Details.}
We use the MinkowskiNet~\cite{minkunet} re-implemented by PCSeg~\cite{pcseg} as our baseline. Our TASeg is trained with the SGD~\cite{sgd} optimizer on 4 A100 GPUs with batch size 6 for 12 and 36 epochs on SemanticKITTI and nuScenes datasets. The learning rate and weight decay are set to 0.02 and 0.0001. The window size of temporal point clouds is set to 16. Due to a large redundancy of images, we use a step of 12 and a window size of 48 for temporal images.
Our data augmentation strategy includes random flipping, rotation, scaling, transformation, LaserMix~\cite{lasermix} and PolarMix~\cite{polarmix}.
During the inference, the LiDAR semantic predictions and image features of the previous timestamp can be saved, so there is no time cost for generating historical predictions and processing temporal image aggregation.
More details are provided in the supplementary material.

\subsection{Comparison with State-of-the-art Methods}
We compare our TASeg with state-of-the-art LiDAR segmentation methods on SemanticKITTI single-scan and multi-scan track, and nuScenes LiDAR semantic segmentation track, as summarized in Table \ref{tab:kitti_test}, \ref{tab:kitti_multi_test} and \ref{tab:nuscenes_test}.
On the SemanticKITTI single-scan track, our TASeg is 1.3 mIoU higher than UniSeg~\cite{uniseg}, the best-performing published algorithm on the SemanticKITTI single-scan leaderboard.
On the SemanticKITTI multi-scan track, our approach outperforms the best multi-scan method 2DPASS~\cite{2dpass} by 3.3 mIoU.
For nuScenes LiDAR semantic segmentation, we also achieve superior performance over previous algorithms. As shown in Table~\ref{tab:nuscenes_test}, the proposed approach surpasses the UniSeg~\cite{uniseg} by 1.1 mIoU.
In particular, our approach holds the highest entry on all three tracks. These encouraging results strongly show the effectiveness of our approach.

\begin{table*}[t]
    \renewcommand\tabcolsep{1.5pt}
    \begin{center}
        {\fontfamily{cmr}
            \resizebox{0.9\textwidth}{!}{
                \begin{tabular}{lc|cccccccccccccccc}
\hline
Method& 
\rotatebox{90}{mIoU}& \rotatebox{90}{barrier}& \rotatebox{90}{bicycle}& \rotatebox{90}{bus}&
\rotatebox{90}{car}& \rotatebox{90}{construction~}& \rotatebox{90}{motorcycle}& \rotatebox{90}{pedestrian}&
\rotatebox{90}{traffic-cone}& \rotatebox{90}{trailer}& \rotatebox{90}{truck}& \rotatebox{90}{driveable}&
\rotatebox{90}{other-flat}& \rotatebox{90}{sidewalk}& \rotatebox{90}{terrain}& \rotatebox{90}{manmade}&
\rotatebox{90}{vegetation}\\
\hline
\hline
Cylinder3D~\cite{cylinder3d} & 77.2 & 82.8 & 29.8 & 84.3 & 89.4 & 63.0 & 79.3 & 77.2 & 73.4 & 84.6 & 69.1 & 97.7 & 70.2 & 80.3 & 75.5 & 90.4 & 87.6 \\
SPVCNN~\cite{spvnas} & 77.4 & 80.0 & 30.0 & 91.9 & 90.8 & 64.7 & 79.0 & 75.6 & 70.9 & 81.0 & 74.6 & 97.4 & 69.2 & 80.0 & 76.1 & 89.3 & 87.1 \\
2D3DNet~\cite{2d3dnet} & 80.0 & 83.0 & 59.4 & 88.0 & 85.1 & 63.7 & 84.4 & 82.0 & 76.0 & 84.8 & 71.9 & 96.9 & 67.4 & 79.8 & 76.0 & 92.1 & 89.2 \\
2DPASS~\cite{2dpass} & 80.8 & 81.7 & 55.3 & 92.0 & 91.8 & 73.3 & 86.5 & 78.5 & 72.5 & 84.7 & 75.5 & 97.6 & 69.1 & 79.9 & 75.5 & 90.2 & 88.0 \\
LidarMultiNet~\cite{lidarmultinet} & 81.4 & 80.4 & 48.4 & 94.3 & 90.0 & 71.5 & 87.2 & 85.2 & 80.4 & 86.9 & 74.8 & 97.8 & 67.3 & 80.7 & 76.5 & 92.1 & 89.6 \\
 MSeg3D~\cite{mseg3d} & 81.1 & 83.1 & 42.5 & 94.9 & 92.0 & 67.1 & 78.6 & 85.7 & 80.5 & 87.5 & 77.3 & 97.7 & 69.8 & 81.2 & 77.8 & 92.4 & 90.1 \\
SphereFormer~\cite{sphereformer} & 81.9 & 83.3 & 39.2 & 94.7 & 92.5 & 77.5 & 84.2 & 84.4 & 79.1 & 88.4 & 78.3 & 97.9 & 69.0 & 81.5 & 77.2 & 93.4 & 90.2 \\
UniSeg~\cite{uniseg} & 83.5 & 85.9 & 71.2 & 92.1 & 91.6 & 80.5 & 88.0 & 80.9 & 76.0 & 86.3 & 76.7 & 97.7 & 71.8 & 80.7 & 76.7 &  91.3 & 88.8 \\
\hline
% TASeg (Ours) & \textbf{84.1} & 87.0 & 67.0 & 89.9 & 92.1 & 78.5 & 88.5 & 86.1 & 81.8 & 88.4 & 75.6 & 97.7 & 70.7 & 80.7 & 77.9 & 93.3 & 91.0 \\
TASeg (Ours) & \textbf{84.6} & 87.1 & 69.4 & 90.5 & 92.2 & 78.7 & 90.4 & 86.3 & 81.9 & 88.3 & 75.9 & 97.8 & 70.9 & 81.0 & 78.2 & 93.4 & 91.2 \\
\hline
    \end{tabular}
        }}
    \end{center}
    \vspace{-4mm}
    \caption{Performance comparison with start-of-the-art methods on \textit{nuScenes} test set.}
    \vspace{-4mm}
    \label{tab:nuscenes_test}
\end{table*}

\vspace{-2mm}
\subsection{Ablation studies}
We examine the effect of each component through detailed ablations. 
Unless otherwise specified, the following experiments are trained on the training set of the SemanticKITTI single-scan task and evaluated on the validation set.

\paragraph{Effect of Flexible Step Aggregation.}
In Table \ref{tab:kitti_fsa}, we compare our FSA with other multi-frame algorithms.
MSeg3D~\cite{mseg3d} utilizes temporal point clouds by directly concatenating all points, which introduces a large memory consumption and achieves limited improvement. SVQNet~\cite{svqnet} leverages cross-attention to fuse temporal features while it can only handle fewer LiDAR scans. When increasing LiDAR scans, the memory overhead is also huge. 
Our FSA assigns different classes with different temporal steps, which can leverage long-term temporal information with less memory and time costs. Moreover, thanks to the utilization of historical priors, our FSA achieves better results (71.3 mIoU) than directly concatenating (69.9 mIoU), given the same temporal window size.
To explore the effect of different window sizes on FSA, we present Table \ref{tab:window_size_fsa}.
Results show that the performance is saturated with a window size of 24.

\vspace{-3mm}
\paragraph{Effect of Group Division.}
To investigate the effect of different group divisions for FSA, we provide Table \ref{tab:kitti_division}.
For \textit{division1}, we simply divide all classes into three groups according to their performance on the val set.
The \textit{group1} consists of the classes in [90, 100) mIoU, such as cars and roads.
The \textit{group2} consists of classes in [80, 90) mIoU, such as motorcycles. The \textit{group3} contains the remaining classes. 
We assign \textit{group1}, \textit{group2} 
 and \textit{group3} with the step of $\infty$, 4 and 2. 
A step of $\infty$ means we do not aggregate temporal points for classes in the group due to their near-saturate performances.
The result shows that \textit{division1} can reduce the time and memory costs largely compared to directly stacking (the $2^{st}$ row in Table \ref{tab:kitti_fsa}). 
We can also use performance \textit{top1-6}, \textit{top6-12}, and \textit{top12-19} for division, resulting in \textit{division2}. 
The group division is robust as long as it follows the principle that more difficult classes need smaller steps. 
To further reduce the memory consumption, we can finetune the group division, such as moving large-size classes (e.g. other-ground and terrain) in \textit{group3} to \textit{group2} (\textit{division3}) or moving large-size classes in \textit{group3} and \textit{group2} to a new group with step of 8 (\textit{division4}). Besides, we can divide each class group into a close and a distant group according to the distance (such as 30 m). Considering close areas need fewer temporal points, we can use twice the original step for temporal aggregation of close groups (\textit{division5}).

\begin{table}[t]
    \footnotesize 
    \vspace{1mm}
    \begin{center}
    \resizebox{0.94\columnwidth}{!}{
        \begin{tabular}{l|c|c|c}
            \hline
            Method&~Latency~&~Memory~&~mIoU\\
            \hline
            \hline
            Baseline                    &63ms  &5.0G  &68.9 \\\hline
            + MSeg3D\dag   &284ms &69.7G &69.9 \\
            + SVQNet\dag   &121ms  &15.4G &69.5 \\
            + FSA                        &79ms  &8.5G  &71.3 \\\hline
            + FSA w/ gt mask              &79ms  &8.5G  &75.9\\
            + FSA + M.D.                   &79ms  &8.5G  &71.8\\
            + FSA + M.D. $\times0.75$      &72ms  &6.9G  & 71.6\\
            \hline
            \end{tabular}
        }
    \end{center}
    \vspace{-5mm}
    \caption{Comparison with different multi-frame algorithms. \dag~represents our re-implementation. ``M.D.'' represents Mask Distillation. ``$\times75$'' represents that we reduce the model parameters to 75\%. }
    \vspace{-2mm}
    \label{tab:kitti_fsa}
\end{table}

\begin{table}[t]
    \renewcommand{\tabcolsep}{3.5pt}
    \begin{center}
        {\fontfamily{cmr}
            \resizebox{0.45\textwidth}{!}{
                \begin{tabular}{c|c|c|c|c|c|c|c}
                    \hline
                    Window size & 4 & 8 & 12 & 16 & 20 & 24 & 28\\
                    \hline
                    \hline
                    mIoU & 69.9	& 70.4 & 70.9 & 71.3 & 71.4 & \textbf{71.6} & 71.4\\
                    \hline
                \end{tabular}
                    }
        }
    \end{center}
    \vspace{-5mm}
    \caption{Ablation on different window sizes for FSA.}
    \label{tab:window_size_fsa}
    \vspace{-2mm}
\end{table}

\begin{table}[t]
    \footnotesize 
    \begin{center}
    \resizebox{0.9\columnwidth}{!}{
        \begin{tabular}{l|c|c|c}
            \hline
            Method&~Latency~&~Memory~&~mIoU\\
            \hline
            \hline 
    FSA w/ division1   &92ms  &14.1G  &70.8 \\
    FSA w/ division2   &83ms  &7.2G   &70.6 \\
    FSA w/ division3   &79ms  &8.5G   &71.3 \\
    FSA w/ division4   &74ms  &7.7G   &71.2 \\
    FSA w/ division5   &74ms  &7.4G   &71.2 \\
            \hline
            \end{tabular}
        }
    \end{center}
    \vspace{-5mm}
    \caption{Ablation on different group divisions for FSA.}
    \vspace{-6mm}
    \label{tab:kitti_division}
\end{table}

\vspace{-4mm}
\paragraph{Effect of Mask Distillation.}
At the $5^{th}$ row of Table \ref{tab:kitti_fsa}, we use gt masks of historical frames for FSA. The performance is greatly boosted. 
Considering we cannot get ground truth at inference, we use it as a teacher to distill the model trained with pseudo masks. 
The result shows that the student is improved by 0.5 mIoU after distillation. 
With 75\% model complexity, we can further reduce the time and memory cost and still achieve a leading performance.

\vspace{-2mm}
\paragraph{Effect of Temporal Image Aggregation and Fusion.}
TIAF delivers a temporal multi-modal fusion scheme to
make full use of temporal images. 
As shown in Table \ref{tab:tiaf}, with zero image, only TLAD is used, achieving 71.8 mIoU.
With one image, only present images are utilized.
Due to the limited FOV of the present camera, only part of the LiDAR points can gather image features, 
which limits the multi-modal fusion.
Our TIAF leverages temporal images to enlarge the FOV of the camera and enhance present features. The results show that with the number of images increasing, the performance rises gradually, which verifies the effectiveness of our TIAF.
When using seven temporal images, we achieve 1.0 mIoU improvement on the strong baseline.
Note that there is no extra cost to generate historical image features because we can save them at the past moments.

\vspace{-2mm}
\begin{table}[h]
    \renewcommand{\tabcolsep}{6.5pt}
    \begin{center}
    \resizebox{0.45\textwidth}{!}{
        \begin{tabular}{c|c|c|c|c|c}
            \hline
            Num of images & 0 & 1 & 3 & 5 & 7\\
            \hline\hline
            mIoU & 71.8 & 72.1 & 72.4 & 72.7 & \textbf{72.8}\\\hline
        \end{tabular}
    }
    \end{center}
    \vspace{-5mm}
    \caption{Ablation on the number of temporal images for TIAF.}
    \vspace{-7mm}
    \label{tab:tiaf}
\end{table}

\begin{table}[h]
\renewcommand\tabcolsep{1.5pt}
    \begin{center}
    \resizebox{0.33\textwidth}{!}{
        \begin{tabular}{c|ccc|c}
            \hline
            Exp &2D Sup. & 3D Sup. & Multi-Scale & mIoU\\
            \hline
            \hline
            (a) &          &\checkmark &\checkmark & 72.4 \\
            (b) &\checkmark &          &\checkmark & 72.5 \\
            (c) &\checkmark &\checkmark &          & 72.4 \\
            (d) &\checkmark &\checkmark &\checkmark & \textbf{72.7} \\\hline
        \end{tabular}
    }
    \end{center}
    \vspace{-4mm}
    \caption{Effect of different supervisions and multi-scale for TIAF.}
    \vspace{-2mm}
    \label{tab:supervision}
\end{table}

\vspace{-2mm}
We also provide the ablation on different supervisions and multi-scale features for TIAF in Table \ref{tab:supervision}.
In TIAF, we leverage 2D and 3D supervision to guide the extracted image features to be more conducive for segmentation. Multi-scale is also used to provide more discriminative features.
The result shows that each of the designs is beneficial for final performance.
Note that our temporal multi-modal method is orthogonal to other single-frame multi-modal methods. We use a simple pixel-to-point mapping at the feature projection stage. More complex methods \cite{mseg3d, uniseg} can also be utilized, but it is not the focus of our TIAF.

\vspace{-5mm}
\paragraph{Effect of Static-Moving Switch Augmentation.}
To verify our SMSA, we present Table \ref{tab:kitti_smsa}.
From the results, we can find that without SMSA, the accuracy of the model on many classes is lower than 10 IoU.
After using SMSA, the performance of moving truck, moving other-vehicle and static bicyclist classes is improved by more than 20 IoU, and the overall performance boosts from 61.3 mIoU to 65.7 mIoU, which strongly demonstrates the efficacy of our SMSA.

\vspace{-3mm}
\begin{table}[h]
    \renewcommand\tabcolsep{1.5pt} 
    \begin{center}
        {\fontfamily{cmr}
            \resizebox{0.47\textwidth}{!}{%
                \begin{tabular}{c|c|cccccccccccc}
                    \hline
                    SMSA &
                    \rotatebox{90}{mIoU}& \rotatebox{90}{car-s}& \rotatebox{90}{car-m}&
                    \rotatebox{90}{truc.-s}& \rotatebox{90}{truc.-m}& \rotatebox{90}{othe.-s}& \rotatebox{90}{othe.-m~}&
                    \rotatebox{90}{pers.-s}& \rotatebox{90}{pers.-m}& \rotatebox{90}{bicy.-s}& \rotatebox{90}{bicy.-m}&
                    \rotatebox{90}{moto.-s}& \rotatebox{90}{moto.-m~} \\
                    \hline 
                    \hline
     & 61.3 & 95.6 & 79.3 & 49.2 & 9.1 & 61.0 & 5.7 & 39.0 & 87.0 & 0.0 & 72.6 & 4.3 & 76.4  \\
    \checkmark & \textbf{65.7} & 94.8 & 72.8     & 53.6 & 35.2    & 71.4 & 53.2    & 36.7 & 80.4    & 32.4 & 75.2    & 12.3 & 89.6 \\
                    \hline
        \end{tabular}}}
    \end{center}
    \vspace{-5mm}
    \caption{Ablation study of our SMSA on the \textit{SemanticKITTI} test benchmark of the multi-scan track.}
    \vspace{-6mm}
    \label{tab:kitti_smsa}
\end{table}

\paragraph{Generalization to More Architectures.} In addition to MinkowskiNet, we also validate our temporal aggregation method on Cylinder3D~\cite{cylinder3d} and SPVNAS~\cite{spvnas}. Detailed results are summarized in Table~\ref{tab:kitti_general}. Our TASeg brings impressive improvements to both baselines, which demonstrates the strong generalization ability of our method.

\begin{table}[ht]
    \footnotesize 
    \begin{center}
    \resizebox{0.68\columnwidth}{!}{
        \begin{tabular}{l|c}
            \hline
            Method & ~mIoU\\
            \hline
            \hline
            Cylinder3D~\cite{cylinder3d}\dag &66.0 \\
            Cylinder3D~\cite{cylinder3d}\dag + TASeg  &\textbf{69.6} \\
            \hline
            SPVNAS~\cite{spvnas}\dag  & 68.6 \\
            SPVNAS~\cite{spvnas}\dag + TASeg  & \textbf{72.7} \\
            \hline
            \end{tabular}
        }
    \end{center}
    \vspace{-4mm}
    \caption{Performance of applying our TASeg to different baselines. \dag~represents our re-implementation.}
    \vspace{-3mm}
    \label{tab:kitti_general}
\end{table}

\vspace{-5mm}
\paragraph{Comparison on Complexity, Latency and Accuracy.}
As shown in Table \ref{tab:kitti_latency}, when not utilizing temporal images, 
our TASeg achieves superior performance than other methods with comparable or less complexity and latency.
With temporal images, TASeg can maintain lower complexity and latency than UniSeg~\cite{uniseg} while achieving higher accuracy.
Besides, with 75\% parameters, the latency of TASeg can be further reduced, and the accuracy is not affected much. 

\begin{table}[ht]
    \footnotesize 
    \begin{center}
    \vspace{-1mm}
    \resizebox{0.98\columnwidth}{!}{
        \begin{tabular}{l|c|c|c|c}
            \hline
            Method & Input & Params &~Latency~&~mIoU\\
            \hline
            \hline
            MinkowskiNet\dag & L &37.9M &63ms  &68.4 \\
            Cylinder3D\dag & L &55.9M &75ms  &66.0 \\
            SPVNAS\dag & L &96.7M &105ms  &68.6 \\
            RPVNet\dag & L &119.0M &110ms  &68.9 \\
            UniSeg & L+C & 147.6M &145ms &71.3 \\\hline
            TASeg wo/ TIAF & L+T   & 37.9M &79ms  & 71.8\\
            TASeg         & L+C+T & 46.7M &116ms  & \textbf{72.7}\\
            TASeg $\times0.75$ & L+C+T & 27.4M & 108ms & 72.5 \\
            \hline
        \end{tabular}
        }
    \end{center}
    \vspace{-4mm}
    \caption{Comparison of the model complexity, latency and accuracy on SemanticKITTI \textit{val} set. 
    \dag~represents our re-implementation on NVIDIA A100 for a fair comparison. ``L'',``C'' and ``T'' represent LiDAR, Camera and Temporal data, respectively. }
    \vspace{-3mm}
    \label{tab:kitti_latency}
\end{table}

\vspace{-1mm}
\section{Conclusion}\label{conclusion}
\vspace{-1mm}
In this paper, we present TASeg to fully exploit temporal point clouds and temporal images. 
Our TASeg is comprised of TLAD, TIAF and SMSA. 
We perform extensive experiments to validate the efficacy of each component. Moreover, our TASeg set a new state-of-the-art on three challenging tracks, demonstrating the superiority of our method.
% , \textit{i.e.}, SemanticKITTI and nuScenes.
% Our approach achieves appealing results and brings significant improvement to different backbones.
% on SemanticKITTI single-scan track, multi-scan track and nuScenes LiDAR segmentation track,
% Specially, our TASeg sets a new state-of-the-art three challenging tracks,  which strongly demonstrates the efficacy of the proposed method. 

\section{Acknowledgements}
This work was supported in part by The National Nature Science Foundation of China (Grant Nos: 62273301, 62273302, 62036009, 61936006, 62273303), in part by Yongjiang Talent Introduction Programme (Grant No: 2022A-240-G, 2023A-197-G), 
in part by the National Key R\&D Program of China (NO.2022ZD0160100). This work was done during the first author's internship at Shanghai AI Laboratory.

\clearpage
%%%%%%%%% REFERENCES
{\small
\bibliographystyle{ieeenat_fullname}
\bibliography{egbib}

\begin{thebibliography}{47}
\providecommand{\natexlab}[1]{#1}
\providecommand{\url}[1]{\texttt{#1}}
\expandafter\ifx\csname urlstyle\endcsname\relax
  \providecommand{\doi}[1]{doi: #1}\else
  \providecommand{\doi}{doi: \begingroup \urlstyle{rm}\Url}\fi

\bibitem[Aygun et~al.(2021)Aygun, Osep, Weber, Maximov, Stachniss, Behley, and Leal-Taix{\'e}]{4d-pls}
Mehmet Aygun, Aljosa Osep, Mark Weber, Maxim Maximov, Cyrill Stachniss, Jens Behley, and Laura Leal-Taix{\'e}.
\newblock 4d panoptic lidar segmentation.
\newblock In \emph{Proceedings of the IEEE/CVF Conference on Computer Vision and Pattern Recognition}, pages 5527--5537, 2021.

\bibitem[Cao et~al.(2020)Cao, Lu, Lu, Pang, Liu, and Yuille]{ASAP-Net}
Hanwen Cao, Yongyi Lu, Cewu Lu, Bo Pang, Gongshen Liu, and Alan Yuille.
\newblock Asap-net: Attention and structure aware point cloud sequence segmentation.
\newblock \emph{arXiv preprint arXiv:2008.05149}, 2020.

\bibitem[Chen et~al.(2023{\natexlab{a}})Chen, Wu, Chitta, Jaeger, Geiger, and Li]{chen2023end}
Li Chen, Penghao Wu, Kashyap Chitta, Bernhard Jaeger, Andreas Geiger, and Hongyang Li.
\newblock End-to-end autonomous driving: Challenges and frontiers.
\newblock \emph{arXiv preprint arXiv:2306.16927}, 2023{\natexlab{a}}.

\bibitem[Chen et~al.(2023{\natexlab{b}})Chen, Xu, Zou, Cao, Yeung, and Fang]{svqnet}
Xuechao Chen, Shuangjie Xu, Xiaoyi Zou, Tongyi Cao, Dit-Yan Yeung, and Lu Fang.
\newblock Svqnet: Sparse voxel-adjacent query network for 4d spatio-temporal lidar semantic segmentation.
\newblock In \emph{Proceedings of the IEEE/CVF International Conference on Computer Vision}, pages 8569--8578, 2023{\natexlab{b}}.

\bibitem[Cheng et~al.(2021)Cheng, Razani, Taghavi, Li, and Liu]{af2s3net}
Ran Cheng, Ryan Razani, Ehsan Taghavi, Enxu Li, and Bingbing Liu.
\newblock (af)2-{S3N}et: Attentive {F}eature {F}usion with {A}daptive {F}eature {S}election for {S}parse {S}emantic {S}egmentation {N}etwork.
\newblock In \emph{IEEE Conference on Computer Vision and Pattern Recognition}, pages 12547--12556, 2021.

\bibitem[Choy et~al.(2019{\natexlab{a}})Choy, Gwak, and Savarese]{4d_minknet}
Christopher Choy, JunYoung Gwak, and Silvio Savarese.
\newblock 4d spatio-temporal convnets: Minkowski convolutional neural networks.
\newblock In \emph{CVPR}, 2019{\natexlab{a}}.

\bibitem[Choy et~al.(2019{\natexlab{b}})Choy, Gwak, and Savarese]{minkunet}
Christopher Choy, JunYoung Gwak, and Silvio Savarese.
\newblock 4{D} {S}patio-{T}emporal {C}onv{N}ets: {M}inkowski {C}onvolutional {N}eural {N}etworks.
\newblock In \emph{IEEE/CVF Conference on Computer Vision and Pattern Recognition}, pages 3075--3084, 2019{\natexlab{b}}.

\bibitem[Dewan and Burgard(2020)]{deeptemporalseg}
Ayush Dewan and Wolfram Burgard.
\newblock Deeptemporalseg: Temporally consistent semantic segmentation of 3d lidar scans.
\newblock In \emph{2020 IEEE International Conference on Robotics and Automation (ICRA)}, pages 2624--2630. IEEE, 2020.

\bibitem[Duerr et~al.(2020)Duerr, Pfaller, Weigel, and Beyerer]{temporallidarseg}
Fabian Duerr, Mario Pfaller, Hendrik Weigel, and J{\"u}rgen Beyerer.
\newblock Lidar-based recurrent 3d semantic segmentation with temporal memory alignment.
\newblock In \emph{2020 International Conference on 3D Vision (3DV)}, pages 781--790. IEEE, 2020.

\bibitem[El~Madawi et~al.(2019)El~Madawi, Rashed, El~Sallab, Nasr, Kamel, and Yogamani]{el2019rgb}
Khaled El~Madawi, Hazem Rashed, Ahmad El~Sallab, Omar Nasr, Hanan Kamel, and Senthil Yogamani.
\newblock Rgb and lidar fusion based 3d semantic segmentation for autonomous driving.
\newblock In \emph{IEEE Intelligent Transportation Systems Conference}, pages 7--12, 2019.

\bibitem[Genova et~al.(2021)Genova, Yin, Kundu, Pantofaru, Cole, Sud, Brewington, Shucker, and Funkhouser]{2d3dnet}
Kyle Genova, Xiaoqi Yin, Abhijit Kundu, Caroline Pantofaru, Forrester Cole, Avneesh Sud, Brian Brewington, Brian Shucker, and Thomas Funkhouser.
\newblock Learning 3d semantic segmentation with only 2d image supervision.
\newblock In \emph{3DV}, 2021.

\bibitem[Hinton et~al.(2015)Hinton, Vinyals, and Dean]{kd}
Geoffrey Hinton, Oriol Vinyals, and Jeff Dean.
\newblock Distilling the knowledge in a neural network.
\newblock \emph{arXiv preprint arXiv:1503.02531}, 2015.

\bibitem[Hou et~al.(2022)Hou, Zhu, Ma, Loy, and Li]{pvkd}
Yuenan Hou, Xinge Zhu, Yuexin Ma, Chen~Change Loy, and Yikang Li.
\newblock Point-to-{V}oxel {K}nowledge {D}istillation for {L}i{DAR} {S}emantic {S}egmentation.
\newblock In \emph{IEEE Conference on Computer Vision and Pattern Recognition}, pages 8479--8488, 2022.

\bibitem[Hu et~al.(2020)Hu, Yang, Xie, Rosa, Guo, Wang, Trigoni, and Markham]{randla-net}
Qingyong Hu, Bo Yang, Linhai Xie, Stefano Rosa, Yulan Guo, Zhihua Wang, Niki Trigoni, and Andrew Markham.
\newblock Randla-net: Efficient semantic segmentation of large-scale point clouds.
\newblock \emph{Proceedings of the IEEE Conference on Computer Vision and Pattern Recognition}, 2020.

\bibitem[Kong et~al.(2023{\natexlab{a}})Kong, Liu, Chen, Ma, Zhu, Li, Hou, Qiao, and Liu]{rangeformer}
Lingdong Kong, Youquan Liu, Runnan Chen, Yuexin Ma, Xinge Zhu, Yikang Li, Yuenan Hou, Yu Qiao, and Ziwei Liu.
\newblock Rethinking range view representation for lidar segmentation.
\newblock In \emph{Proceedings of the IEEE/CVF International Conference on Computer Vision}, pages 228--240, 2023{\natexlab{a}}.

\bibitem[Kong et~al.(2023{\natexlab{b}})Kong, Ren, Pan, and Liu]{lasermix}
Lingdong Kong, Jiawei Ren, Liang Pan, and Ziwei Liu.
\newblock Lasermix for semi-supervised lidar semantic segmentation.
\newblock In \emph{Proceedings of the IEEE/CVF Conference on Computer Vision and Pattern Recognition}, pages 21705--21715, 2023{\natexlab{b}}.

\bibitem[Kreuzberg et~al.(2022)Kreuzberg, Zulfikar, Mahadevan, Engelmann, and Leibe]{4d-stop}
Lars Kreuzberg, Idil~Esen Zulfikar, Sabarinath Mahadevan, Francis Engelmann, and Bastian Leibe.
\newblock 4d-stop: Panoptic segmentation of 4d lidar using spatio-temporal object proposal generation and aggregation.
\newblock In \emph{European Conference on Computer Vision}, pages 537--553. Springer, 2022.

\bibitem[Krispel et~al.(2020)Krispel, Opitz, Waltner, Possegger, and Bischof]{fuseseg}
Georg Krispel, Michael Opitz, Georg Waltner, Horst Possegger, and Horst Bischof.
\newblock Fuseseg: Lidar point cloud segmentation fusing multi-modal data.
\newblock In \emph{IEEE/CVF Winter Conference on Applications of Computer Vision}, pages 1874--1883, 2020.

\bibitem[Lai et~al.(2023)Lai, Chen, Lu, Liu, and Jia]{sphereformer}
Xin Lai, Yukang Chen, Fanbin Lu, Jianhui Liu, and Jiaya Jia.
\newblock Spherical transformer for lidar-based 3d recognition.
\newblock \emph{arXiv preprint arXiv:2303.12766}, 2023.

\bibitem[Li et~al.(2023{\natexlab{a}})Li, Dai, Han, and Ding]{mseg3d}
Jiale Li, Hang Dai, Hao Han, and Yong Ding.
\newblock M{S}eg3{D}: Multi-modal 3{D} {S}emantic {S}egmentation for {A}utonomous {D}riving.
\newblock \emph{arXiv preprint arXiv:2303.08600}, 2023{\natexlab{a}}.

\bibitem[Li et~al.(2023{\natexlab{b}})Li, Chen, Liu, and Jiang]{depthformer}
Zhenyu Li, Zehui Chen, Xianming Liu, and Junjun Jiang.
\newblock Depthformer: Exploiting long-range correlation and local information for accurate monocular depth estimation.
\newblock \emph{Machine Intelligence Research}, 20\penalty0 (6):\penalty0 837--854, 2023{\natexlab{b}}.

\bibitem[Liu et~al.(2023{\natexlab{a}})Liu, Chang, Liu, Wu, Ma, and Qi]{mars3d}
Jiahui Liu, Chirui Chang, Jianhui Liu, Xiaoyang Wu, Lan Ma, and Xiaojuan Qi.
\newblock Mars3d: A plug-and-play motion-aware model for semantic segmentation on multi-scan 3d point clouds.
\newblock In \emph{Proceedings of the IEEE/CVF Conference on Computer Vision and Pattern Recognition}, pages 9372--9381, 2023{\natexlab{a}}.

\bibitem[Liu et~al.(2023{\natexlab{b}})Liu, Bai, Kong, Chen, Hou, Shi, and Li]{pcseg}
Youquan Liu, Yeqi Bai, Lingdong Kong, Runnan Chen, Yuenan Hou, Botian Shi, and Yikang Li.
\newblock Pcseg: An open source point cloud segmentation codebase.
\newblock \url{https://github.com/PJLab-ADG/PCSeg}, 2023{\natexlab{b}}.

\bibitem[Liu et~al.(2023{\natexlab{c}})Liu, Chen, Li, Kong, Yang, Xia, Bai, Zhu, Ma, Li, et~al.]{uniseg}
Youquan Liu, Runnan Chen, Xin Li, Lingdong Kong, Yuchen Yang, Zhaoyang Xia, Yeqi Bai, Xinge Zhu, Yuexin Ma, Yikang Li, et~al.
\newblock Uniseg: A unified multi-modal lidar segmentation network and the openpcseg codebase.
\newblock In \emph{Proceedings of the IEEE/CVF International Conference on Computer Vision}, pages 21662--21673, 2023{\natexlab{c}}.

\bibitem[Milioto et~al.(2019)Milioto, Vizzo, Behley, and Stachniss]{rangenet++}
Andres Milioto, Ignacio Vizzo, Jens Behley, and Cyrill Stachniss.
\newblock Rangenet++: Fast and accurate lidar semantic segmentation.
\newblock In \emph{Proc. of the IEEE/RSJ Intl. Conf. on Intelligent Robots and Systems (IROS)}, 2019.

\bibitem[Qi et~al.(2017{\natexlab{a}})Qi, Su, Mo, and Guibas]{pointnet}
Charles~R Qi, Hao Su, Kaichun Mo, and Leonidas~J Guibas.
\newblock Pointnet: Deep learning on point sets for 3d classification and segmentation.
\newblock In \emph{Proceedings of the IEEE conference on computer vision and pattern recognition}, pages 652--660, 2017{\natexlab{a}}.

\bibitem[Qi et~al.(2017{\natexlab{b}})Qi, Yi, Su, and Guibas]{pointnet++}
Charles~Ruizhongtai Qi, Li Yi, Hao Su, and Leonidas~J Guibas.
\newblock Pointnet++: Deep hierarchical feature learning on point sets in a metric space.
\newblock In \emph{Advances in neural information processing systems}, pages 5099--5108, 2017{\natexlab{b}}.

\bibitem[Robbins and Monro(1951)]{sgd}
Herbert Robbins and Sutton Monro.
\newblock A {S}tochastic {A}pproximation {M}ethod.
\newblock \emph{The annals of mathematical statistics}, pages 400--407, 1951.

\bibitem[Rosu et~al.(2019)Rosu, Sch{\"u}tt, Quenzel, and Behnke]{latticenet}
Radu~Alexandru Rosu, Peer Sch{\"u}tt, Jan Quenzel, and Sven Behnke.
\newblock Latticenet: Fast point cloud segmentation using permutohedral lattices.
\newblock \emph{arXiv preprint arXiv:1912.05905}, 2019.

\bibitem[Shi et~al.(2020)Shi, Lin, Wang, Hung, and Wang]{SpSequenceNet}
Hanyu Shi, Guosheng Lin, Hao Wang, Tzu-Yi Hung, and Zhenhua Wang.
\newblock Spsequencenet: Semantic segmentation network on 4d point clouds.
\newblock In \emph{CVPR}, 2020.

\bibitem[Tang et~al.(2020)Tang, Liu, Zhao, Lin, Lin, Wang, and Han]{spvnas}
Haotian Tang, Zhijian Liu, Shengyu Zhao, Yujun Lin, Ji Lin, Hanrui Wang, and Song Han.
\newblock Searching {E}fficient 3{D} {A}rchitectures with {S}parse {P}oint-{V}oxel {C}onvolution.
\newblock In \emph{European Conference on Computer Vision}, pages 685--702. Springer, 2020.

\bibitem[Tang et~al.(2019)Tang, Folkesson, and Jensfelt]{sparse2dense}
Jiexiong Tang, John Folkesson, and Patric Jensfelt.
\newblock Sparse2dense: From direct sparse odometry to dense 3-d reconstruction.
\newblock \emph{IEEE Robotics and Automation Letters}, 4\penalty0 (2):\penalty0 530--537, 2019.

\bibitem[Wu et~al.(2019)Wu, Zhou, Zhao, Yue, and Keutzer]{squeezesegv2}
Bichen Wu, Xuanyu Zhou, Sicheng Zhao, Xiangyu Yue, and Kurt Keutzer.
\newblock Squeezesegv2: Improved model structure and unsupervised domain adaptation for road-object segmentation from a lidar point cloud.
\newblock In \emph{2019 International Conference on Robotics and Automation (ICRA)}, pages 4376--4382. IEEE, 2019.

\bibitem[Wu et~al.(2022)Wu, Liao, Zhang, Wang, Bai, Cheng, and Liu]{yolop}
Dong Wu, Man-Wen Liao, Wei-Tian Zhang, Xing-Gang Wang, Xiang Bai, Wen-Qing Cheng, and Wen-Yu Liu.
\newblock Yolop: You only look once for panoptic driving perception.
\newblock \emph{Machine Intelligence Research}, 19\penalty0 (6):\penalty0 550--562, 2022.

\bibitem[Xiao et~al.(2022)Xiao, Huang, Guan, Cui, Lu, and Shao]{polarmix}
Aoran Xiao, Jiaxing Huang, Dayan Guan, Kaiwen Cui, Shijian Lu, and Ling Shao.
\newblock Polar{M}ix: A {G}eneral {D}ata {A}ugmentation {T}echnique for {L}i{DAR} {P}oint {C}louds.
\newblock \emph{arXiv preprint arXiv:2208.00223}, 2022.

\bibitem[Xu et~al.(2020)Xu, Wu, Wang, Zhan, Vajda, Keutzer, and Tomizuka]{squeezesegv3}
Chenfeng Xu, Bichen Wu, Zining Wang, Wei Zhan, Peter Vajda, Kurt Keutzer, and Masayoshi Tomizuka.
\newblock Squeezesegv3: Spatially-adaptive convolution for efficient point-cloud segmentation.
\newblock In \emph{European Conference on Computer Vision}, pages 1--19. Springer, 2020.

\bibitem[Xu et~al.(2021)Xu, Zhang, Dou, Zhu, Sun, and Pu]{rpvnet}
Jianyun Xu, Ruixiang Zhang, Jian Dou, Yushi Zhu, Jie Sun, and Shiliang Pu.
\newblock {RPV}net: A {D}eep and {E}fficient {R}ange-{P}oint-{V}oxel {F}usion {N}etwork for {L}idar {P}oint {C}loud {S}egmentation.
\newblock In \emph{IEEE International Conference on Computer Vision}, pages 16024--16033, 2021.

\bibitem[Yan et~al.(2021)Yan, Gao, Li, Zhang, Li, Huang, and Cui]{js3c-net}
Xu Yan, Jiantao Gao, Jie Li, Ruimao Zhang, Zhen Li, Rui Huang, and Shuguang Cui.
\newblock Sparse single sweep lidar point cloud segmentation via learning contextual shape priors from scene completion.
\newblock In \emph{Proceedings of the AAAI Conference on Artificial Intelligence}, pages 3101--3109, 2021.

\bibitem[Yan et~al.(2022)Yan, Gao, Zheng, Zheng, Zhang, Cui, and Li]{2dpass}
Xu Yan, Jiantao Gao, Chaoda Zheng, Chao Zheng, Ruimao Zhang, Shuguang Cui, and Zhen Li.
\newblock 2{DPASS}: 2{D} {P}riors {A}ssisted {S}emantic {S}egmentation on {L}i{DAR} {P}oint {C}louds.
\newblock In \emph{European Conference on Computer Vision}, 2022.

\bibitem[Yang et~al.(2022)Yang, Shi, Ding, Wang, and Qi]{sparsekd}
Jihan Yang, Shaoshuai Shi, Runyu Ding, Zhe Wang, and Xiaojuan Qi.
\newblock Towards efficient 3d object detection with knowledge distillation.
\newblock \emph{Advances in Neural Information Processing Systems}, 35:\penalty0 21300--21313, 2022.

\bibitem[Ye et~al.(2022)Ye, Zhou, Chen, Xie, Wang, Wang, and Foroosh]{lidarmultinet}
Dongqiangzi Ye, Zixiang Zhou, Weijia Chen, Yufei Xie, Yu Wang, Panqu Wang, and Hassan Foroosh.
\newblock Lidarmultinet: Towards a unified multi-task network for lidar perception.
\newblock \emph{arXiv preprint arXiv:2209.09385}, 2022.

\bibitem[Yurtsever et~al.(2020)Yurtsever, Lambert, Carballo, and Takeda]{yurtsever2020survey}
Ekim Yurtsever, Jacob Lambert, Alexander Carballo, and Kazuya Takeda.
\newblock A survey of autonomous driving: Common practices and emerging technologies.
\newblock \emph{IEEE access}, 8:\penalty0 58443--58469, 2020.

\bibitem[Zhang et~al.(2023)Zhang, Zhang, Yu, Yi, Xie, and Ma]{lcps}
Zhiwei Zhang, Zhizhong Zhang, Qian Yu, Ran Yi, Yuan Xie, and Lizhuang Ma.
\newblock Lidar-camera panoptic segmentation via geometry-consistent and semantic-aware alignment.
\newblock In \emph{Proceedings of the IEEE/CVF International Conference on Computer Vision}, pages 3662--3671, 2023.

\bibitem[Zheng et~al.(2022)Zheng, Jiang, Lu, Ye, and Fu]{smf-ssd}
Wu Zheng, Li Jiang, Fanbin Lu, Yangyang Ye, and Chi-Wing Fu.
\newblock Boosting single-frame 3d object detection by simulating multi-frame point clouds.
\newblock In \emph{Proceedings of the 30th ACM International Conference on Multimedia}, pages 4848--4856, 2022.

\bibitem[Zhu et~al.(2023)Zhu, Han, Cai, Borse, Ghaffari, and Porikli]{eq-4d}
Minghan Zhu, Shizhong Han, Hong Cai, Shubhankar Borse, Maani Ghaffari, and Fatih Porikli.
\newblock 4d panoptic segmentation as invariant and equivariant field prediction.
\newblock In \emph{Proceedings of the IEEE/CVF International Conference on Computer Vision (ICCV)}, pages 22488--22498, 2023.

\bibitem[Zhu et~al.(2021)Zhu, Zhou, Wang, Hong, Ma, Li, Li, and Lin]{cylinder3d}
Xinge Zhu, Hui Zhou, Tai Wang, Fangzhou Hong, Yuexin Ma, Wei Li, Hongsheng Li, and Dahua Lin.
\newblock Cylindrical and {A}symmetrical 3{D} {C}onvolution {N}etworks for {L}idar {S}egmentation.
\newblock In \emph{IEEE Conference on Computer Vision and Pattern Recognition}, pages 9939--9948, 2021.

\bibitem[Zhuang et~al.(2021)Zhuang, Li, Jia, Wang, Li, and Tan]{pmf}
Zhuangwei Zhuang, Rong Li, Kui Jia, Qicheng Wang, Yuanqing Li, and Mingkui Tan.
\newblock Perception-aware multi-sensor fusion for 3d lidar semantic segmentation.
\newblock In \emph{ICCV}, 2021.

\end{thebibliography}
}

\end{document}